\documentclass[sigconf, screen]{acmart}
\usepackage{graphicx}
\usepackage{array}
\usepackage{tabularray}
\UseTblrLibrary{diagbox}
\usepackage{multirow}
\AtBeginDocument{%
  }

\setcopyright{acmlicensed}
\copyrightyear{2025}
\acmYear{2025}
\acmDOI{XXXXXXX.XXXXXXX}
\acmConference[MM '25]{Make sure to enter the correct
  conference title from your rights confirmation email}{27-31 October,
  2025}{Dublin, Ireland}
\acmISBN{978-1-4503-XXXX-X/2018/06}

\acmSubmissionID{2708}

\begin{document}

\title{BrainSegDMIF: A Dynamic Fusion-enhanced SAM for Brain Lesion Segmentation}


\author{Hongming Wang}
\authornote{Equal contribution.}
\affiliation{%
  \institution{Southern University of Science and Technology}
  \country{Shenzhen, China}
 }
\email{12333530@mail.sustech.edu.cn}
\orcid{0009-0008-5589-6461}

\author{Yifeng Wu}
\authornotemark[1]
\affiliation{%
  \institution{Southern University of Science and Technology}
  \state{Shenzhen}
  \country{China}
}
\email{yf.wu1@siat.ac.cn}
\orcid{0009-0005-0218-9091}

\author{Huimin Huang}
\affiliation{%
  \institution{Jarvis Research Center, Tencent YouTu Lab}
  \country{Shenzhen, China}
}
\email{huiminhuang@tencent.com}
\orcid{0000-0001-8987-350X}

\author{Hongtao Wu}
\affiliation{%
  \institution{Westlake University}
  \country{Hangzhou, China}
 }
\email{hwu375@connect.hkust-gz.edu.cn}
\orcid{0009-0007-4863-5119}
  
\author{Jiaxuan Jiang}
\affiliation{%
  \institution{Lanzhou University $\&$ Westlake University}
  \city{Lanzhou}
  \country{China}}
\email{jiangjx2023@lzu.edu.cn}
\orcid{0009-0001-9686-4836}

\author{Xiaodong Zhang}
\affiliation{%
  \institution{Shenzhen University of Advanced Technology}
  \city{Shenzhen}
  \country{China}}
\email{zhangxd0530@gmail.com}
\orcid{0000-0002-1665-8187}

\author{Hao Zheng}
\affiliation{%
  \institution{Tencent YouTu Lab}
  \city{Shenzhen}
  \country{China}}
\email{howzheng@tencent.com}
\orcid{0000-0001-7193-6242}

\author{Yawen Huang}
\affiliation{%
  \institution{Tencent YouTu Lab}
  \city{Shenzhen}
  \country{China}}
\email{yawenhuang@tencent.com}
\orcid{0000-0002-9569-269X}

\author{Xian Wu}
\affiliation{%
  \institution{Tencent YouTu Lab}
  \city{Shenzhen}
  \country{China}}
\email{kevinxwu@tencent.com}
\orcid{0000-0003-1118-9710}

\author{Yefeng Zheng}
\authornotemark[2] 
\affiliation{%
  \institution{Westlake University}
  \city{Hangzhou}
  \country{China}}
\email{zhengyefeng@westlake.edu.cn}
\orcid{0000-0003-2195-2847}

\author{Jinping Xu}
\authornotemark[3] 
\affiliation{%
  \institution{Shenzhen Institutes of Advanced Technology, Chinese Academy of Sciences}
  \city{Shenzhen}
  \country{China}}
\email{jp.xu@siat.ac.cn}
\orcid{0000-0003-3877-5445}

\author{Jing Cheng}
\authornote{Corresponding author}
\affiliation{%
  \institution{Shenzhen Institutes of Advanced Technology, Chinese Academy of Sciences}
  \city{Shenzhen}
  \country{China}}
\email{jing.cheng@siat.ac.cn}
\orcid{0000-0001-9098-8048}


\renewcommand{\shortauthors}{Hongming Wang et al.}

\begin{abstract}
  The segmentation of substantial brain lesions is a significant and challenging task in the field of medical image segmentation. Substantial brain lesions in brain imaging exhibit high heterogeneity, with indistinct boundaries between lesion regions and normal brain tissue. Small lesions in single slices are difficult to identify, making the accurate and reproducible segmentation of abnormal regions, as well as their feature description, highly complex. Existing methods have the following limitations: 1) They rely solely on single-modal information for learning, neglecting the multi-modal information commonly used in diagnosis. This hampers the ability to comprehensively acquire brain lesion information from multiple perspectives and prevents the effective integration and utilization of multi-modal data inputs, thereby limiting a holistic understanding of lesions. 2) They are constrained by the amount of data available, leading to low sensitivity to small lesions and difficulty in detecting subtle pathological changes. 3) Current SAM-based models rely on external prompts, which cannot achieve automatic segmentation and, to some extent, affect diagnostic efficiency.To address these issues, we have developed a large-scale fully automated segmentation model specifically designed for brain lesion segmentation, named BrainSegDMIF. This model has the following features: 1) Dynamic Modal Interactive Fusion (DMIF) module that processes and integrates multi-modal data during the encoding process, providing the SAM encoder with more comprehensive modal information. 2) Layer-by-Layer Upsampling Decoder, enabling the model to extract rich low-level and high-level features even with limited data, thereby detecting the presence of small lesions. 3) Automatic segmentation masks, allowing the model to generate lesion masks automatically without requiring manual prompts.We tested and evaluated our model on two common brain disease segmentation benchmarks, including cases of focal cortical dysplasia and gliomas. Our model outperformed existing state-of-the-art methods across four metrics.
  
\end{abstract}





\maketitle
\begin{figure*}[!htb]  
	\centering  
	\includegraphics[width=\textwidth]{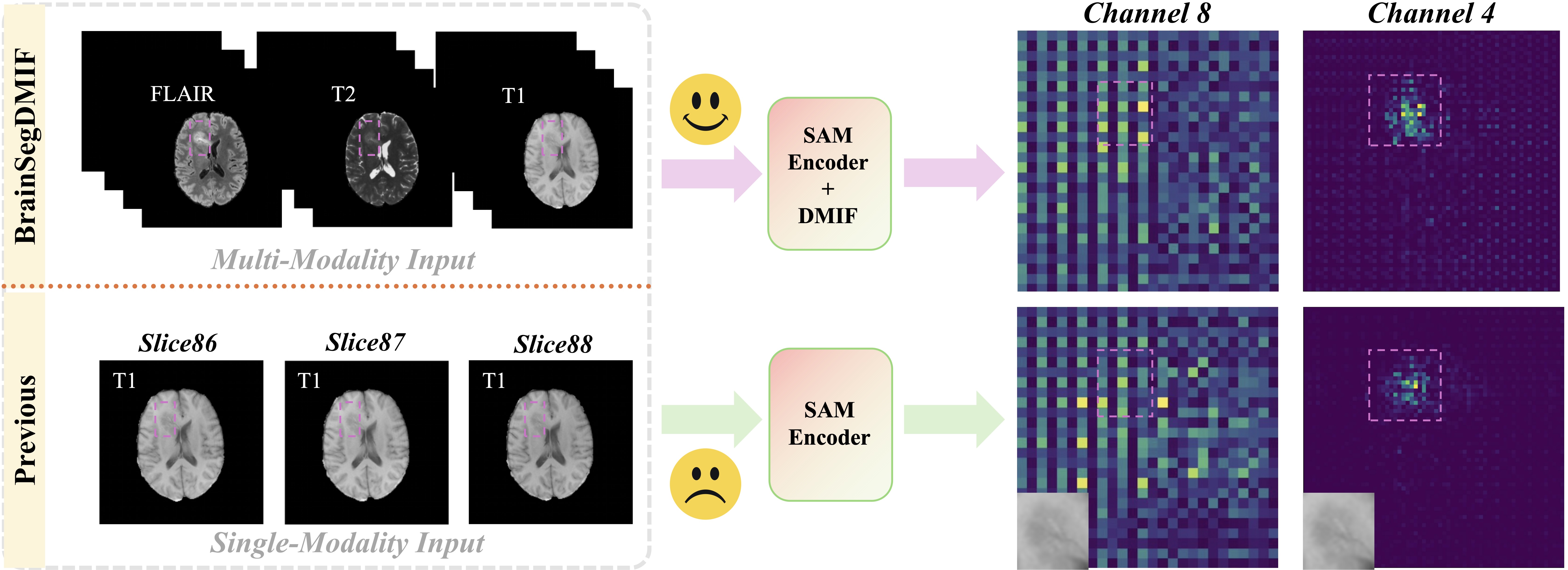}  
	\caption{The figure compares feature extraction using multimodal versus unimodal data.
    The left panel shows multimodal brain lesion images, while the right panel displays lesion features extracted using our methods.
    Features extracted from unimodal data are more dispersed, making it difficult to distinguish between lesion and non-lesion regions. In contrast, after multimodal fusion using the DMIF module, lesion regions appear significantly brighter and features are more concentrated.
    }
	\label{fig:1}  
        \Description{The motivation of our paper.}
\end{figure*}

\section{Introduction}
Brain parenchymal lesions represent one of the most severe and complex challenges in the medical field, encompassing tumors, vascular diseases, trauma, and other conditions that may be congenital or acquired. Segmentation constitutes the initial stage in the treatment planning of brain parenchymal lesions \cite{niyazi2016estro,kruser2019nrg} , playing a crucial role in the diagnosis, treatment, and monitoring of various diseases. This stage primarily relies on physicians manually delineating lesion regions \cite{kruser2019nrg} , which requires substantial time and professional expertise. Due to the heterogeneity of lesions, blurred boundaries, and diverse morphological characteristics, segmentation becomes challenging, resulting in inconsistencies in the size of regions manually segmented by different physicians \cite{jungo2018effect,joskowicz2019inter} . Computer-assisted brain parenchymal lesion segmentation represents a particularly important method that can help hospitals and patients save considerable time in disease detection, improve physician efficiency and treatment success rates, and eliminate issues of segmentation inconsistency.

In recent years, deep learning~\cite{li2024flaws, huang2024p2sam,wu2023mask, wu2024semi,wu2024rainmamba,yuan2024auformer,liu2024multi,gong2024cross,gong2022person}, has been increasingly applied to medical image segmentation \cite{ronneberger2015u,wang2024video,isensee2021nnu,sun2022few,li2024flaws,wang2024pv}. Deep learning models such as U-net and its variants have demonstrated excellent performance and accuracy in medical image segmentation \cite{ronneberger2015u,isensee2019automated,huang2023stu} . Although these convolutional neural network-based methods are effective in extracting local image features, they are limited by their local receptive fields \cite{luo2016understanding},making them unable to utilize global contextual information and process long-range dependencies in image data. Recent studies have proposed Vision Transformers (ViTs), which have made significant progress in addressing global context and long-range dependency problems \cite{raghu2021vision,hatamizadeh2023global}. Based on improvements to ViTs, several highly effective models have emerged, such as SwinUNet \cite{cao2022swin}. Notably, in the field of medical image segmentation, models like UNETR and nnFormer\cite{hatamizadeh2022unetr,zhou2023nnformer} have demonstrated outstanding performance.However, both CNN-based and ViT-based models require large amounts of annotated data, which presents a major challenge in the medical field. Considering the similarities between segmentation tasks, using pre-trained weights from natural image models has become possible.

The Segment Anything Model (SAM) \cite{kirillov2023segment}, developed by Meta AI, consists of a Transformer-based image encoder coupled with a lightweight decoder. As a novel foundational visual segmentation model trained on billions of images, it has shown tremendous potential in medical imaging, particularly in segmenting organs with clear boundaries \cite{cheng2023sam}. However, due to the heterogeneity and blurred boundaries characteristic of lesions, significant challenges exist in lesion segmentation, especially for small lesions. Given SAM's relatively simple decoder structure and the small proportion of minor lesions in images, whose shape, texture, and other features are less distinct compared to normal lesions, SAM may fail to clearly recognize or segment small lesions, particularly with medical image datasets that are typically limited in scale and diversity. SAM lacks multi-modal support, without accounting for multi-modal data input possibilities, resulting in deficiencies in multi-modal data processing and limiting its learning to single-modality data with only simple processing for multi-modal data.

MRI serves as the cornerstone of clinical diagnosis and treatment when evaluating brain lesions. As a high-resolution imaging technology, MRI offers multi-modal imaging capabilities, clearly displaying subtle changes in brain structures. Clinicians typically analyze multiple MRI sequence parameters comprehensively to achieve a thorough assessment of the patient's condition. Common MRI modalities include T1, T2, FLAIR, as shown in Figure \ref{fig:1}. Single-modality MRI has inherent limitations in displaying lesion boundaries and features, making it difficult to fully characterize heterogeneous lesion features \cite{liu2023deep}. Although multi-modal data contains rich complementary information, many current research methods have not fully exploited its potential, with most simply using multi-modal data as direct input to models \cite{liu2022sf}, lacking effective multi-modal information fusion strategies. This simplified processing approach limits the model's ability to learn and integrate lesion information from multiple perspectives, resulting in underutilization of valuable data resources.

To address the above challenges in brain lesion segmentation, we propose BrainSegDMIF—a fully automatic 2D brain parenchymal lesion segmentation model that incorporates multi-modal and multi-scale capabilities based on SAM. The model innovatively devises a multimodal fusion module to effectively integrate and transfer information across different modalities. This module efficiently extracts and integrates multi-modal features, capturing complementary information of the same lesion from different perspectives. To address the model's insufficient sensitivity in recognizing small lesions, we further designed a layer-by-layer upsampling decoder that employs a multi-scale feature fusion strategy, systematically restoring the spatial resolution of deep feature maps extracted during the model's encoding phase while preserving rich semantic information of the image data. We also optimized SAM's functionality to automatically generate lesion masks without prompts. 

Our main contributions can be summarized as follows:
\begin{itemize}
\item In response to SAM's insufficient utilization of multi-modal data, we designed an efficient multi-modal image fusion module (DMIF) that integrates multi-modal image data during the image encoding stage, enhancing the model's image comprehension capabilities.
\item Addressing the heterogeneity, blurred boundaries, and poor visualization of small lesions, we designed a layer-by-layer upsampling decoder that employs a multi-scale information fusion strategy to enhance the model's segmentation accuracy and sensitivity to lesions.
\item By optimizing SAM, our model can automatically generate high-quality image segmentation masks without prompts, eliminating the traditional SAM's dependence on clicking, boxing, or text prompts, thereby improving the model's practicality and efficiency in real-world application scenarios.
\end{itemize}

\section{Related Work}
\subsection{SAM in Medical Image Analysis}
Large foundation models represent one of the most dynamic and rapidly evolving domains in artificial intelligence research. As a novel foundational visual segmentation model, SAM (Segment Anything Model) has demonstrated exceptional unsupervised and zero-shot generalization capabilities through training on the extensive SA-1B dataset.

Despite SAM's significant advances in natural image segmentation, it encounters substantial performance issues when applied to medical image segmentation tasks \cite{deng2023sam,bernard2018deep,ma2024segment}. This performance gap mainly stems from the severe scarcity of medical data in SAM’s training set, in stark contrast to the abundance of natural images.\cite{zhang2023survey}. This data imbalance has prevented SAM from learning sufficient anatomical structure representations in the medical domain, which are crucial for reliable medical image understanding.

Consequently, adapting SAM for medical image segmentation has become a key research focus, with many studies optimizing SAM for medical tasks \cite{wu2025medical,ma2024segment,zhang2024segment}. Researchers mainly use either comprehensive or parameter-efficient fine-tuning. Comprehensive fine-tuning involves thorough parameter adjustments of the pre-trained model, typically updating most or all model parameters to better suit specific tasks or domains. Through comprehensive fine-tuning of vanilla SAM on large medical datasets, researchers have achieved state-of-the-art results \cite{ma2024segment}. However, comprehensive fine-tuning demands substantial memory resources and computational power, although research indicates that transferring pre-trained models to medical imaging tasks is highly feasible \cite{raghu2019transfusion}.

Parameter-efficient fine-tuning methods achieve efficient model adaptation by updating only a small portion of parameters in pre-trained models. Unlike comprehensive fine-tuning, this approach freezes the majority of parameters and only learns an extremely small subset, typically less than 5\% of the total parameter count. Current research predominantly focuses on fine-tuning SAM with specific medical segmentation datasets to adapt it to particular tasks. Wu et al. \cite{wu2025medical}proposed MSA, a straightforward adapter technique that integrates specific medical domain knowledge into SAM by inserting adapters into the original model, thereby enhancing its capabilities in medical tasks. Zhang et al. \cite{zhang2023customized} fine-tuned SAM's encoder component using LoRA while simultaneously fine-tuning the decoder for abdominal segmentation. Integrating SAM into medical image segmentation further involves model modifications and new module designs. Wu et al. \cite{wu2024mamba} enabled SAM to adapt to lesion image segmentation by introducing LoRA and designing new decoder modules, thus addressing SAM's deficiencies in medical lesion recognition.

\subsection{Multimodal medical image segmentation}
Multimodal learning enhances model representation capability by leveraging multi-source information. MRI, as the preferred modality for brain-related diseases, captures tissue characteristics and pathological information through sequences such as T1, T2, FLAIR, and T1CE, providing complementary perspectives.However, existing studies have merely treated the data as prior information in a simplistic manner. For instance, Wang et al.\cite{wang2018automatic} designed a cascaded architecture that combines multimodal data as input for brain tumor segmentation. Myronenko et al.\cite{myronenko20183d} also integrated multimodal data as input, augmenting a U-Net architecture with a VAE branch for image reconstruction. Z. Jiang et al.\cite{jiang2020two} proposed a two-stage cascaded U-Net for segmenting multimodal brain tumor data. Y. Zhang\cite{zhang2021modality} embedded multimodal CT images through a fully connected layer as input. Despite their effectiveness, these approaches fail to explore how to integrate complementary information from different modalities to form a unified and efficient representation.This has motivated us to investigate multimodal fusion in the context of image segmentation.

\begin{figure*}[t]
\includegraphics[width=1\textwidth]{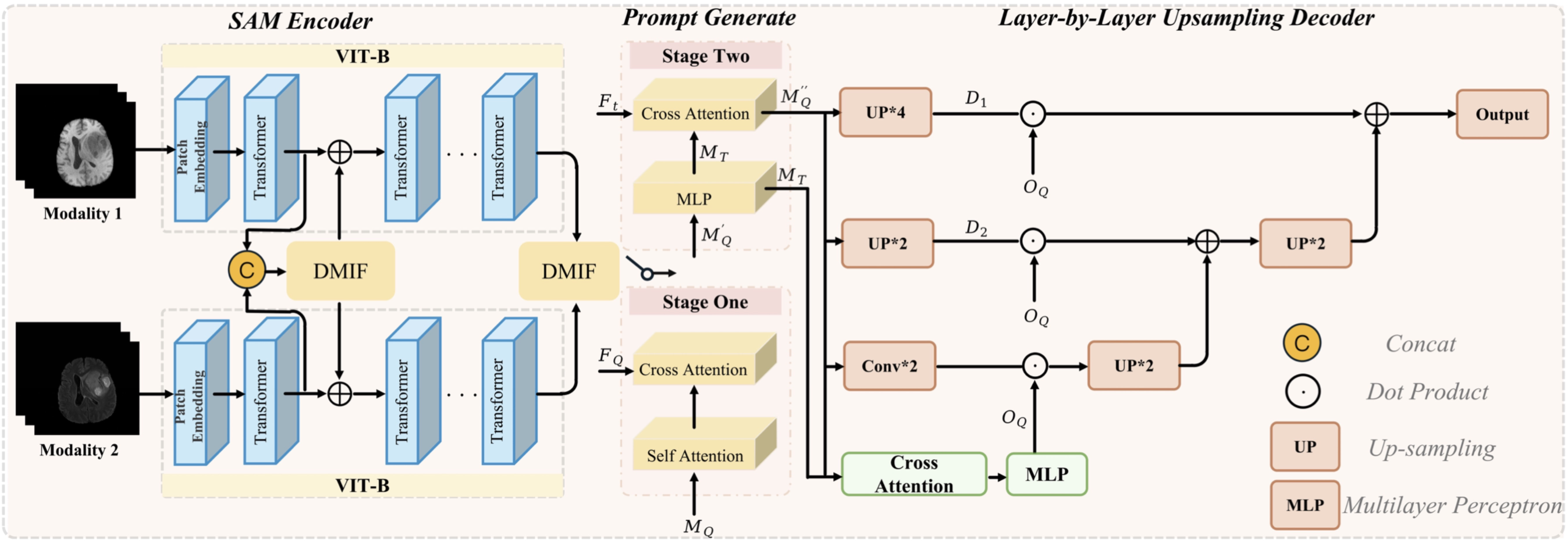}
    \caption{ Schematic of the BrainSegDMIF model. Given a set of multimodal brain-lesion images, the DMIF module first fuses the multimodal features and integrates them with the image features before forwarding the combined representation to the encoder. A decoder with layer-wise refined upsampling then generates the segmentation masks. Image features are merged with mask tokens via an attention mechanism, enabling the capture of critical characteristics within the lesion regions.  }
    \label{fig:method}
\end{figure*}

\section{Methods}
In the context of multi-modal medical image segmentation, we define the training set as $D_M=\{(x_{i}^M,y_{i}^M)\}_{i=1}^{n_M}$ , where $x_{i}^M \in\mathbb{R}^{H\times W\times C}$ represents the i-th multi-modal sample, typically comprising image data from various modalities (such as MRI, CT, etc.), and $y_{i}^M\in\mathbb{R}^{H\times W}$ denotes its corresponding ground truth annotation. The term $n_{M}$ indicates the number of training samples. The objective of this task is to maximize the similarity between the predicted labels $\overline{y}_i^T$ and an unseen multi-modal dataset $D_{T}=\{(x_{i}^{T},y_{i}^{T})\}_{i=1}^{n_{T}}$.

The proposed BrainSegDMIF architecture is illustrated in Figure \ref{fig:method}, comprising three principal components: a modality fusion encoder, an automatic mask generator, and a progressive upsampling decoder. For each patient's multi-modal data $x_{i}^M$, we initially extract features through modality-specific encoders, yielding feature representations $f_{s_1}$ , $f_{s_2}$ , $\ldots$ , $f_{s_m}$ across different modalities, where $m$ denotes the number of modalities. These modal features are subsequently directed to the Dynamic Modality Interaction Fusion (DMIF) module, which adaptively generates fusion weights by integrating multi-modal data, performing weighted fusion of features from different modalities to enhance complementary information while suppressing redundant information. The fused features $f_{fus}$ are transmitted during the encoding phase to each modality's encoder where they are integrated with the data learned by that specific encoder, compensating for learning limitations caused by information deficiencies in individual modalities, thereby enhancing the model's comprehension capability and representational efficacy for multi-modal data. The $f_{fus}$ obtained from the final layer is further propagated to the decoder, where the multi-scale fusion mechanism of the decoder gradually increases the scale of the generated lesion mask, guiding model learning through loss function optimization and consequently improving segmentation accuracy.

\begin{figure*}[t]
\includegraphics[width=1\textwidth]{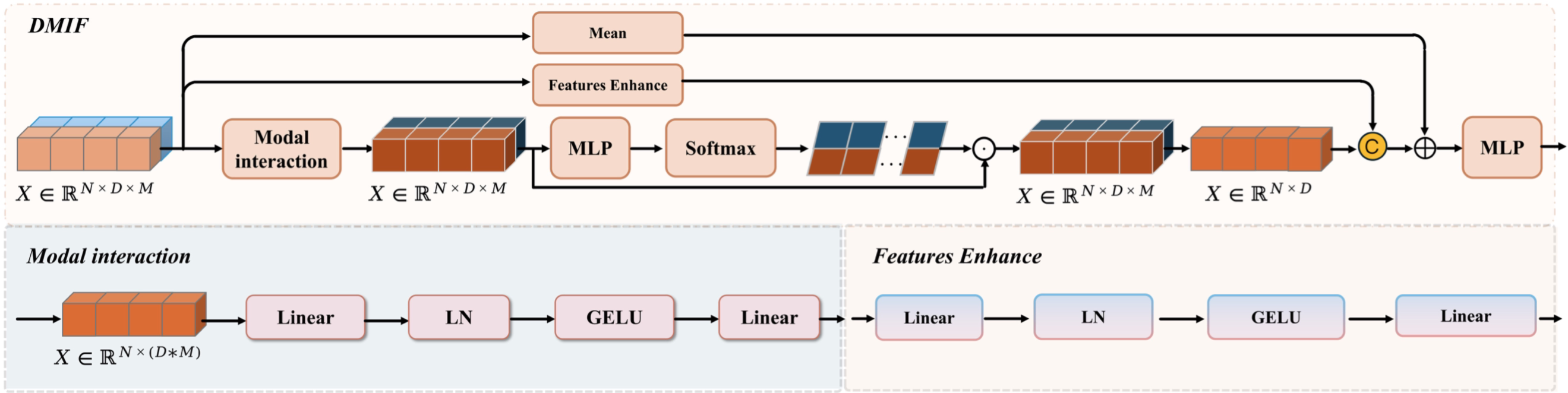}
    \caption{ The figure presents the schematic of our Dynamic Modal Interactive Fusion (DMIF) module. This module accepts concatenated multi-modal image tokens. Through dynamic interactions, it fuses multi-modal information. Dynamic weights are generated to emphasize salient features. The fused features are then refined via these weights. Finally, the aggregated features are enhanced using a broadcasting mechanism with residual connections, yielding the final multi-modal features.}
    \label{fig:DMIF}
\end{figure*}

\subsection{Dynamic Modal Interactive Fusion}
In clinical practice, multimodal imaging is widely used for disease diagnosis, and effectively utilizing multimodal data remains a fundamental challenge in medical image segmentation. To address this, we designed the Dynamic Modal Interactive Fusion (DMIF) module to aggregate medical imaging information from diverse modalities and provide lesion features from multiple perspectives to enhance model performance. The architecture of this module is illustrated in Figure \ref{fig:DMIF}.

The DMIF module consists of three key components: the inter-modal interaction module, the dynamic weight generation module, and the adaptive feature aggregation module. These components interact and fuse data to learn complementary information from multimodal data adaptively, enabling the model to accommodate multi-modal inputs.

Since most clinical multimodal data are image-based, we used a unified SAM Encoder for feature extraction and encoding to avoid inconsistencies in sequence length and feature dimensions caused by different extraction methods. Specifically, for an image $x_{i}^m $ from modality $m$ , we encoded it using multi-layer Vision Transformers (ViTs) to obtain features $f_{s_1}$ , $f_{s_2}$ , $\ldots$ , $f_{s_m}$ at different layers. These features were combined into a high-dimensional feature tensor $f_{i}\in\mathbb{R}^{N \times D\times M}$  where $i$ denotes the i-th layer. This process maps multi-modal data from their original domains into a unified feature space, forming a high-dimensional feature $f_{i}$ that serves as input to the inter-modal interaction module. By unifying features into a shared space, we established a foundation for subsequent semantic alignment and feature interaction.

\subsubsection{Intermodal interaction layer}
The inter-modal interaction layer is designed to enable effective interaction and influence among features from different modalities. Multimodal data often originates from distinct data domains, each with unique characteristics and representations. For instance, T1 modality is more effective in observing gray matter in the brain, while FLAIR focuses on cerebrospinal fluid. These differences create semantic gaps between modalities, making alignment challenging. Despite these differences, multimodal data shares implicit associations as it represents the same underlying subject. To address this complexity, we propose a multimodal feature semantic alignment mechanism. This mechanism leverages a nonlinear transformation module to achieve deep interaction and semantic alignment between modalities, constructing a new unified feature representation space based on $f_{i}$.

We designed a nonlinear transformation module to achieve implicit interaction and semantic alignment between modalities:
\begin{equation}
  f_{i}^{\prime}=\mathcal{F}_{\mathrm{Norm}}\left(\mathbf{W_1}\cdot f_{i}+\mathbf{b_1}\right),
\end{equation}
\begin{equation}
  f_{i}^{\prime\prime}=\mathbf{W_2}\cdot\mathcal{F}_{\mathrm{GELU}}(f_{i}^{\prime})+\mathbf{b_2},
\end{equation}
where $\mathcal{F}_{\mathrm{Norm}}$ denotes layer normalization, which is used for feature normalization to ensure semantic space alignment of features. $\mathcal{F}_{\mathrm{GELU}}$ represents the activation function, capturing complex interactions between different modalities through nonlinear transformation. The second linear transformation remaps features to the original dimension, ensuring that the output features of each modality are influenced by all other modalities. This approach enables the model to interactively integrate features from all modalities adaptively. On the basis of integrating information from other modalities for each individual modality, the original feature information of each modality is retained.

\subsubsection{Dynamic weight generation}
In multimodal learning, substantial discrepancies across modalities in visibility, noise levels, and other factors are often overlooked during feature fusion, ultimately degrading representation quality. To achieve precise modality feature fusion, we introduce an adaptive weight generation mechanism. By generating weights in each encoder component, it adaptively determines each modality's importance in each encoding state. These weights compile multimodal fusion representations in each encoding layer, as shown in Figure \ref{fig:DMIF}. First, the interactively fused multimodal features $f_{i}^{\prime\prime}$ are passed through a multi-layer fully connected network to learn feature weights at different encoding layers. Then, softmax normalization is applied to obtain each feature's percentage among all features. The resulting weights are multiplied with the corresponding feature vectors to obtain a multimodal feature $F_{i} \in \mathbb{R}^{N \times D}$ that adaptively adjusts feature distribution based on current needs. This feature, containing all features of the current encoding layer, effectively suppresses the impact of noise and irrelevant modalities on the model. The process is defined as:
\begin{equation}
  \mathbf{W}_{model} = \sigma(MLP(f_{i}^{\prime\prime})),
\end{equation}
\begin{equation}
  \sigma(x_i) = \frac{e^{x_{i}}}{\sum_{j=1}^M e^{x_{j}}} \ \ \ for\ i=1,2,\dots,M,
\end{equation}
where $\mathbf{W}_{model}$ denotes the learned adaptive multimodal feature weight matrix, and $\sigma(\cdot)$ epresents the softmax function, ensuring the generated weights sum to 1 to reflect each feature's importance.
The generated weight matrix $\mathbf{W}_{model}$ is element-wise multiplied with the interactively fused features:
\begin{equation}
  f_{i}^{\prime\prime\prime} = f_{i}^{\prime\prime} \odot \mathbf{W}_{modal} ,
\end{equation}
where $\odot$ denotes the Hadamard product.

\subsubsection{Feature Convergence}
From the previous step, we obtained multimodal features $f_{i}^{\prime\prime\prime}$ with varying feature distributions at each encoding step. These features can adaptively adjust themselves, but interaction between modalities has not yet been established, as interaction was only completed within the DMIF module. To transmit the interacted features to each modality encoding component, we summed the features obtained from the previous step to form the final fused multimodal feature $F_{i} \in\mathbb{R}^{N \times D} $ :
\begin{equation}
  F_{i} =  \sum_{j=1}^{M}f_{i}^{\prime\prime\prime}[...,j],
\end{equation}
this feature incorporates the importance of each modality at this layer. To maintain feature diversity, we designed parallel feature enhancement branches. These branches use a feature enhancement function $\gamma(\cdot)$ to extract useful information from the originally concatenated features through nonlinear transformations, constructing a representational subspace complementary to the main fusion path. This supplements the main aggregation path, and combining information from both paths yields complete multimodal information $F{c}$ . Finally, to ensure information integrity and gradient flow, we introduced a residual linear structure that uses the average representation of multimodal features:
\begin{equation}
  f_{fus} = F_{c} + \frac{1}{M}\sum_{j=1}^{M}f_{i}[...,j],
\end{equation}
this design ensures fairness of the residual branch in multimodal scenarios, preventing residual information from biasing toward specific modalities and enhancing the diversity of feature representation.

\subsection{Prompt Generate}
Since SAM relies on prompts, we designed a Prompt generate module to enable SAM to perform automatic lesion segmentation. By incorporating attention mechanisms, this module facilitates interaction between image tokens and mask tokens, allowing the generated mask tokens to capture key information related to lesions in the image. This approach enables the automatic learning of prompts associated with lesion regions, thereby achieving automatic segmentation.

The process is divided into two stages. In the first stage, we define the mask token as $M_{Q} \in \mathbb{R}^{N \times D}$ , where $N$  is the number of mask tokens and $D$  is the feature dimension. We employ self-attention to learn the internal contextual information of the mask tokens. The attention-encoded mask tokens are represented as $\tilde{M}_Q$. Given the multimodal features $F_{r}\in \mathbb{R}^{N \times D}$ , we compute the attention between $\tilde{M}_Q$ and $F_{r}$ :
\begin{equation}
  M_Q^{^{\prime}}=\mathrm{Attn}(\tilde{M}_Q,F_r)=\mathrm{Softmax}\left(\frac{\tilde{M}_Q\tilde{W}_{Q}(F_rW_K^{^{r}})^T}{\sqrt{d_k}}\right)F_rW_V^{^{r}},
\end{equation}
this establishes a relationship between the multimodal features and the mask tokens, aligning them semantically and enriching the mask tokens with missing information.

In the second stage, we pass $M_Q^{^{\prime}}$ through a fully connected layer to obtain the final mask features $M_{T}\in \mathbb{R}^{N \times D}$  This enhances the model's expressiveness and aids in more complex processing. To enable the multimodal features to focus on key information in the mask features and capture lesion characteristics, we compute the attention between the multimodal features $F_{t}$ and $M_{T}$ :
\begin{equation}
  M_{Q}^{\prime \prime}=\mathrm{Attn}(F_t,M_T),
\end{equation}
we thereby obtain the enhanced multimodal features $M_{Q}^{\prime \prime} \in \mathbb{R}^{N \times D}$ , which integrate mask information. These features serve as the input to the decoder, providing comprehensive multimodal information for the decoding process.

\subsection{Layer-by-Layer Upsampling Decoder}
After slicing brain lesion volumes, individual slices may contain only tiny lesions. These small lesions are typically characterized by fuzzy boundaries and strong heterogeneity. To address these challenges, we designed a layer-wise upsampling decoder, as depicted in Figure \ref{fig:method}. Our decoder comprises multiple upsampling and convolution modules.

The enhanced features $M_{Q}^{\prime \prime}$ are passed through upsampling modules of different sizes, utilizing 2D transposed convolution operations to obtain information at different resolutions, namely $D_1$ , $D_2$ , and $D_3$:
\begin{equation}
  D_i = \mathcal{F}_{up}^i(M_{Q}^{\prime \prime}) = \text{ConvTranspose2D}(M_{Q}^{\prime \prime}; \theta_i), \quad i \in {1,2,3}
\end{equation}
where $F_{up}^i$ enotes the i-th upsampling module, composed of transposed convolutions with different strides and kernel sizes. $\theta_{i}$ represents the learnable parameters of the module.

We integrate mask information $M_{T}$ and image information $M_{Q}^{\prime \prime}$ via attention to form a comprehensive feature representation $O_{Q}\in \mathbb{R}^{N \times D}$ , which captures interactions between mask and multimodal information. We then use $O_{Q}$ to enhance the previously obtained multi-scale features $D_1$ , $D_2$ , and $D_3$ through feature modulation:
\begin{equation}
  D_i^{enhanced}=G(O_Q)\odot D_i, \quad i \in {1,2,3}
\end{equation}
where $G$ is a feature mapping function that converts $O_Q$ into a representation compatible with $D_i$ , and $\odot$ denotes element-wise multiplication. This operation enables the model to adaptively emphasize key information in multi-scale features based on $O_Q$ .

Finally, we resize $D_1^{enhanced}$ and $D_2^{enhanced}$ to the same spatial dimensions using bilinear interpolation and combine them with $D_3^{enhanced}$ to generate the final mask:
\begin{equation}
D_i^{resized}=I(D_i^{enhanced},s_i), \quad i \in {1,2}
\end{equation}
where $I(\cdot,s_i)$ represents the bilinear interpolation function, and $s_i$ is the scaling factor.

We fuse features across scales progressively:
\begin{equation}
  D_{final}=D_1^{resized}+D_2^{resized}+D_3^{enhanced},
\end{equation}
this multi-scale feature fusion strategy enables our model to leverage spatial information at different resolutions, producing segmentation masks with clear boundaries and complete structures.

\subsubsection{Loss function}
To improve the model's performance in detecting small lesions, we adopted FocalDice Loss as the loss function. This loss function combines Focal Loss and Dice Loss, with a linear weighting ratio of 1:2. The calculation formula is as follows:
\begin{equation}
  FocalDice=\frac{1}{2}Focal+Dice.
\end{equation}

\begin{table*}[]
\caption{Quantitative results of different methods on BraTs2021 and FCD2023 in terms of Dice(\%), IoU(\%), Prec(\%), Sens(\%). * indicates the use of a pre-trained model. MedSAM* and SAM Med2D* utilize their respective ViT-B pre-trained models.}
\begin{tabular}{lclcllllllllll}
\hline
\multicolumn{1}{c}{}                     &                          & \multicolumn{1}{c}{}                       &                                & \multicolumn{4}{c}{BraTS2021}                                                            & \multicolumn{4}{c}{FCD2023}                                       & \multicolumn{2}{c}{Average}     \\
\multicolumn{1}{c}{\multirow{-2}{*}{\#}} & \multirow{-2}{*}{Method} & \multicolumn{1}{c}{\multirow{-2}{*}{Year}} & \multirow{-2}{*}{Params / GF}  & Dice                                  & IoU            & Prec           & Sens           & Dice           & IoU            & Prec           & Sens           & Dice           & IoU            \\ \hline
1                                        & UNETR                    & 2022                                       & 112M / 270                     & 52.26                                 & 48.09          & 56.75          & 49.32          & 26.68          & 16.09          & 33.45          & 25.56          & 39.47          & 32.09          \\
2                                        & SwinUNETR                & 2021                                       & 62M / 214                      & 53.27                                 & 47.14          & 54.43          & 53.33          & 28.64          & 16.09          & 29.71          & 27.86          & 40.96          & 31.62          \\
3                                        & nnFormer                 & 2021                                       & \multicolumn{1}{l}{149M / 119} & 62.36                                 & 54.80          & 63.50          & 61.45          & 43.88          & 30.35          & 46.91          & 45.61          & 53.12          & 42.58          \\
4                                        & MixUNETR                 & 2025                                       & 50M / 228                      & 61.98                                 & 52.14          & 59.65          & 62.31          & 38.61          & 24.52          & 35.56          & 41.72          & 50.30          & 38.33          \\
5                                        & STUNet                   & 2023                                       & 8M / 24                        & 59.80                                 & 51.06          & 60.32          & 57.11          & 27.42          & 16.75          & 28.81          & 26.18          & 43.61          & 33.91          \\
6                                        & SAM 2                    & 2024                                       & 44M  / 128                     & 00.03                                 & 00.01          & 00.02          & 00.02          & 00.03          & 00.01          & 00.01          & 00.02          & 00.03          & 00.01          \\
7                                        & MedSAM*                  & 2024                                       & 70M / 743                      & 68.44                                 & 55.08          & 69.60          & 68.21          & 54.02          & 39.66          & 53.20          & 55.60          & 61.23          & 47.37          \\
8                                        & MedSAM                   & 2024                                       & 70M / 743                      & 71.68                                 & 61.96          & 74.35          & 67.88          & 56.43          & 40.77          & 51.20          & 58.61          & 64.06          & 51.37          \\
9                                        & SAM Med2D*               & 2024                                       & 271M / 130                     & 70.89                                 & 58.95          & 72.51          & 66.28          & 55.83          & 45.25          & 57.31          & 53.17          & 63.36          & 52.10          \\
10                                       & SAM Med2D                & 2024                                       & 271M / 130                     & 72.39                                 & 63.21          & 70.79          & 75.68          & 57.51          & 42.96          & 56.31          & 58.91          & 64.95          & 53.09          \\
11                                       & BrainSegDMIF             & 2025                                       & 354M / 311                     & {\color[HTML]{000000} \textbf{79.64}} & \textbf{68.55} & \textbf{81.88} & \textbf{78.24} & \textbf{64.87} & \textbf{51.07} & \textbf{63.79} & \textbf{68.07} & \textbf{72.26} & \textbf{59.81} \\ \hline
\end{tabular}
\label{table1}
\end{table*}

\begin{figure*}[t]
\includegraphics[width=1\textwidth]{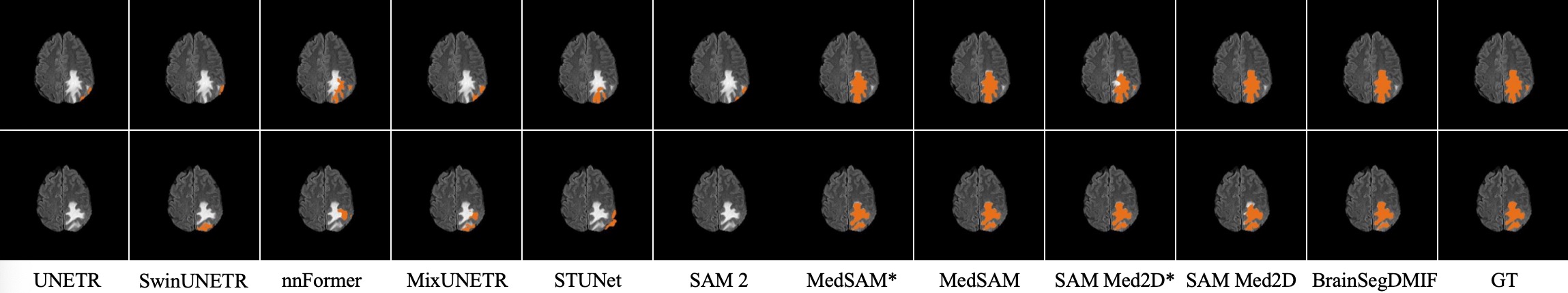}
    \caption{The visualization of our model's segmentation performance on the BraTs21 dataset is presented. * indicates the use of a pre-trained model.}
    \label{fig:result1}
\end{figure*}

\begin{figure*}[t]
\includegraphics[width=1\textwidth]{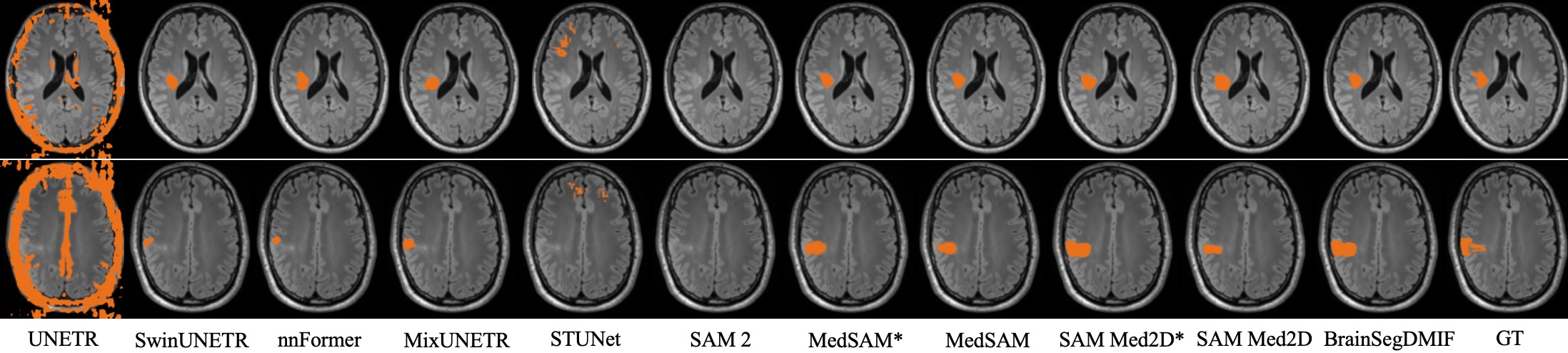}
    \caption{The visualization of segmentation results for various models on the FCD2023 dataset is presented. * indicates the use of a pre-trained model.}
    \label{fig:result2}
\end{figure*}

\section{Experiments}
\subsection{Datasets and Evaluation Metrics}
\subsubsection{datasets}
To evaluate the efficacy of our methodology, we employed two publicly available multi-sequence MRI datasets for segmentation tasks and ablation experiments. The datasets are as follows:

BraTS21 (Brain Tumor Segmentation Challenge)\cite{baid2021rsna}. This dataset represents a comprehensive, publicly available multi-modal brain glioma segmentation dataset, encompassing four MRI modalities: T1, T1CE, T2, and FLAIR. Each modality has dimensions of 240x240x155 (LxWxH). All labels and data have undergone preprocessing, including alignment with a standardized anatomical template, adjustment to uniform resolution (1 mm³), and skull-stripping.

FCD2023\cite{schuch2023open}. This public dataset focuses on Focal Cortical Dysplasia (FCD) and includes three modalities: T1, T2, and FLAIR, from 85 patients. Data acquisition utilized a 32-channel head coil, with image dimensions of 256x256 and voxel size of 1.0 mm×1.0 mm×1.0 mm.

Regarding image preprocessing, our model is designed for 2D image segmentation. We initially performed resampling and alignment of the multi-modal datasets, followed by slicing the 3D data along the third dimension and saving the resulting 2D slices. The provided mask data were adjusted to values within the [0-255] range.
\subsubsection{Evaluation Metrics}
Given that our model focuses on lesion segmentation, to objectively evaluate our model's performance, we selected four of the most commonly used metrics, Dice and IoU (Intersection over Union), as well as Precision and Recall, to fairly evaluate the performance of our model by comparing the final segmentation results. Dice measures the similarity between predicted results and ground-truth labels, while IoU represents the ratio of the intersection to the union of predicted results and ground-truth labels. Precision reflects the accuracy of the model in identifying true positives among all predicted positives, and Recall reflects the model's sensitivity to detecting true positives.

\subsection{Implementation Details}
Our method was implemented using the PyTorch deep learning framework and trained for 200 epochs on four NVIDIA A100 GPUs with 80GB of memory each. We employed the Adam optimizer with an initial learning rate of \(1 \times 10^{-4}\) and utilized a MultiStepLR learning rate scheduler to dynamically adjust the optimizer's learning rate. Specifically, the learning rate was multiplied by 0.5 at the 7th and 12th epochs to gradually reduce it, aiding model convergence. We fine-tuned the SAM-B base model in this work. During training, all images were resized to 256x256 resolution. If an image's width or height was smaller than 256, its borders were padded with black; otherwise, bilinear interpolation was used for resizing.

\subsection{Comparisons With Other Methods}
In this section, we conducted a quantitative comparison with several state-of-the-art methods, including UNETR\cite{hatamizadeh2022unetr}, SwinUNETR\cite{hatamizadeh2021swin}, nnFormer\cite{zhou2023nnformer}, MixUNETR\cite{shen2025mixunetr}, STUNet~\cite{huang2023stu}, SAM 2~\cite{ravi2024sam}, MedSAM~\cite{ma2024segment}, and SAMMed2D\cite{cheng2023sam}. To ensure a fair comparison, the comparison models used the FLAIR modality data, which showed the best segmentation results in Table \ref{table:data}, and were trained according to their default training protocols. The quantitative comparison results are presented in Table \ref{table1}, where our method demonstrates superior performance across all datasets. Specifically, on the BraTS2021 dataset, our method achieved a Dice score of 79.64, an IoU of 68.55, a Precision of 81.88, and a Sensitivity of 78.24, outperforming the second-best method, SAM Med2D, by 7.25, 5.34, 11.09, and 2.56 points, respectively. On the FCD 2023 dataset, our model achieved a Dice score of 64.84, an IoU of 51.07, a Precision of 63.79, and a Sensitivity of 68.07, surpassing SAM Med2D by 7.36, 8.11, 7.48, and 9.16 points, respectively. Compared to MedSAM, which also leverages SAM, our method achieved a Dice score that was 7.96 higher, an IoU that was 6.59 higher, a Precision that was 7.53 higher, and a Sensitivity that was 10.36 higher on the BraTS2021 dataset. The superior segmentation performance of our method can be attributed to modal interaction during training, enabling the model to learn domain knowledge from multimodal data and capture data from different perspectives. Figure \ref{fig:result1} shows the segmentation results of various methods on the BraTS21 dataset, illustrating that our method better identifies lesion regions. Figure \ref{fig:result2} presents the segmentation results of each method on the FCD2023 dataset, where our method more effectively distinguishes between lesion and non-lesion regions.

\begin{table}[]
\caption{The Impact of Multimodal Data on Segmentation}
\label{table:data}
\begin{tabular}{llll|llll}
\hline
\multicolumn{4}{c|}{\textit{Modalities Input}}                                                                                             & \multicolumn{4}{c}{\textit{Metric}}                               \\ \hline
\multicolumn{1}{c}{T1}           & T2                               & T1CE                             & FLAIR                             & \textit{DICE}  & \textit{IoU}   & \textit{Prec}  & \textit{Sens}  \\ \hline
\multicolumn{1}{c}{$\checkmark$} & \multicolumn{1}{c}{$\checkmark$} &                                  &                                   & 73.84          & 64.27          & 70.29          & 76.53          \\
\multicolumn{1}{c}{$\checkmark$} &                                  & \multicolumn{1}{c}{$\checkmark$} &                                   & 72.33          & 62.89          & 74.61          & 71.10          \\
\multicolumn{1}{c}{$\checkmark$} &                                  &                                  & \multicolumn{1}{c|}{$\checkmark$} & 72.58          & 63.56          & 73.12          & 72.19          \\
                                 & $\checkmark$                     & $\checkmark$                     &                                   & 73.91          & 64.33          & 74.97          & 72.13          \\
                                 & $\checkmark$                     &                                  & $\checkmark$                      & 74.16          & 61.36          & 75.34          & 73.79          \\
$\checkmark$                     & $\checkmark$                     & $\checkmark$                     &                                   & 75.12          & 67.01          & 71.35          & \textbf{78.97} \\
$\checkmark$                     & $\checkmark$                     &                                  & $\checkmark$                      & 74.56          & 65.78          & 80.43          & 70.38          \\
$\checkmark$                     &                                  & $\checkmark$                     & $\checkmark$                      & 74.86          & 65.92          & 73.34          & 75.82          \\
\multicolumn{1}{c}{}             & $\checkmark$                     & $\checkmark$                     & $\checkmark$                      & 76.89          & 67.64          & 75.35          & 77.44          \\
$\checkmark$                     & $\checkmark$                     & $\checkmark$                     & $\checkmark$                      & \textbf{79.64} & \textbf{68.55} & \textbf{81.78} & 78.24          \\ \hline
\end{tabular}
\end{table}

\begin{table}[]
\caption{The Effectiveness of DMIF, PG, and LUD, where DMIF stands for Dynamic Modal Interactive Fusion, PG refers to Prompt Generate, and LUD represents Layer-by-Layer Upsampling Decoder.}
\label{table:DMIF}
\begin{tabular}{clc|llll}
\hline
DMIF                             & PG                  & \multicolumn{1}{l|}{LUD} & \textit{DICE}  & \textit{IoU}   & \textit{Prec}  & \textit{Sens}  \\ \hline
$\checkmark$                     & \multicolumn{1}{c}{}             & \multicolumn{1}{l|}{}                                  & 74.53          & 63.71          & 76.29          & 70.26          \\
                                 & \multicolumn{1}{c}{$\checkmark$} &                                                        & 71.29          & 58.45          & 72.77          & 67.94          \\
                                 &                                  & $\checkmark$                                           & 73.93          & 60.28          & 71.08          & 75.34          \\
$\checkmark$                     & \multicolumn{1}{c}{$\checkmark$} & \multicolumn{1}{l|}{}                                  & 75.80          & 65.91          & 79.05          & 73.67          \\
$\checkmark$                     &                                  & $\checkmark$                                           & 76.75          & 65.88          & 80.51          & 75.29          \\
\multicolumn{1}{l}{}             & $\checkmark$                     & $\checkmark$                                           & 74.36          & 64.47          & 73.53          & 75.41          \\
\multicolumn{1}{l}{$\checkmark$} & $\checkmark$                     & $\checkmark$                                           & \textbf{79.64} & \textbf{68.55} & \textbf{81.78} & \textbf{78.24} \\ \hline
\end{tabular}
\end{table}

\subsection{Ablation experiment}
\subsubsection{The Impact of Multimodal Data on Segmentation}
We conducted ablation studies on each modality's impact on the model, as shown in Table \ref{table:data}. From the table, it is evident that using all modalities for segmentation yields the highest Dice, IoU, and Prec scores. When using three modalities, the combination of T2, T1CE, and FLAIR performs the best. However, adding a fourth modality (FLAIR) leads to a lower Sens score compared to using three modalities. This confirms our assertion that not all information in each modality is beneficial for model learning.
\subsubsection{The Effectiveness of DMIF, PG, and LUD}
We conducted paired experiments to analyze the impact of our proposed modules on segmentation results. The model with only the PG module served as the baseline. From Table \ref{table:DMIF}, it can be observed that the model without the DMIF module performed the worst across all four metrics. When only the PG module was used, the model achieved the lowest performance. After adding the LUD module, the Dice score improved by 3.07, the IoU metric increased by 6.02, the Prec metric rose by 0.76, and the Sens metric improved by 7.47. When the DMIF module was added, the model achieved the best performance. This demonstrates that the modules proposed in this study effectively enhance the segmentation performance of the model.

\section{Conclusion}
Our work proposes a novel network named BrainSegDMIF, based on SAM, which explores brain parenchymal lesion segmentation through multimodal fusion and layer-wise upsampling decoding. BrainSegDMIF introduces three key contributions: the multimodal fusion module (DMIF), a layer-wise upsampling decoder, and automatic image segmentation. The multimodal fusion module is integrated into the SAM Encoder, enabling it to learn features from multiple modalities during image encoding by interacting with multimodal data.

Given the potential presence of small lesions in brain parenchymal diseases and the possibility of lesions appearing extremely small in image slices, we designed a layer-wise upsampling decoder. This decoder progressively enlarges feature scales and fuses multi-scale features, enhancing the model's sensitivity to small lesions and improving lesion segmentation accuracy. Additionally, by designing a prompt generator, we achieved fully automatic lesion segmentation.

Experimental results on the BraTS21 and FCD 2023 datasets demonstrate that our network effectively integrates multimodal data, comprehensively learns data features, and achieves superior segmentation accuracy.

\clearpage

\begin{acks}
This work was supported by 
Zhejiang Leading Innovative and Entrepreneur Team Introduction Program (2024R01007).
\end{acks}


\bibliographystyle{ACM-Reference-Format}
\bibliography{sample-base}

\end{document}